\def\year{2021}\relax
\documentclass[letterpaper]{article} % DO NOT CHANGE THIS
\usepackage{aaai21}  % DO NOT CHANGE THIS
\usepackage{times}  % DO NOT CHANGE THIS
\usepackage{helvet} % DO NOT CHANGE THIS
\usepackage{courier}  % DO NOT CHANGE THIS
\usepackage[hyphens]{url}  % DO NOT CHANGE THIS
\usepackage{graphicx} % DO NOT CHANGE THIS
\usepackage{booktabs}
\usepackage{multirow}
\usepackage{subcaption}
\usepackage{color}\usepackage{xcolor}
\usepackage{natbib}  % DO NOT CHANGE THIS AND DO NOT ADD ANY OPTIONS TO IT
\usepackage{caption} % DO NOT CHANGE THIS AND DO NOT ADD ANY OPTIONS TO IT
\newcommand{\alg}{SMART }

\usepackage{soul}
\usepackage[switch]{lineno} 

\newcommand{\specialcell}[2][c]{%
  \begin{tabular}[#1]{@{}c@{}}#2\end{tabular}}

\title{\alg Frame Selection for Action Recognition }
\author{

      Shreyank N Gowda$^1$
   \hspace{1cm}
   Marcus Rohrbach$^2$
   \hspace{1cm}
   Laura Sevilla-Lara$^1$ 
  
 }
 \affiliations{
 $^1$University of Edinburgh \hspace{1cm} $^2$Facebook AI Research
 }
\begin{document}
%\linenumbers
\maketitle

\begin{abstract}
Action recognition is computationally expensive.  In this paper, we address the problem of frame selection to improve the accuracy of action recognition. In particular, we show that selecting good frames helps in action recognition performance even in the trimmed videos domain. Recent work has successfully leveraged frame selection for long, untrimmed videos, where much of the content is not relevant, and easy to discard. In this work, however, we focus on the more standard short, trimmed action recognition problem. We argue that good frame selection can not only reduce the computational cost of action recognition but also increase the accuracy by getting rid of frames that are hard to classify. In contrast to previous work, we propose a method that instead of selecting frames by considering one at a time, considers them jointly. This results in a more efficient selection, where “good" frames are more effectively distributed over the video, like snapshots that tell a story. We call the proposed frame selection SMART and we test it in combination with different backbone architectures and on multiple benchmarks (Kinetics, Something-something, UCF101). We show that the SMART frame selection consistently improves the accuracy compared to other frame selection strategies while reducing the computational cost by a factor of 4 to 10 times. Additionally, we show that when the primary goal is recognition performance, our selection strategy can improve over recent state-of-the-art models and frame selection strategies on various benchmarks (UCF101, HMDB51, FCVID, and ActivityNet).
\end{abstract}

\section{Introduction}
Video processing is computationally expensive. At the same time, the amount of video content being generated is increasing fast and constitutes a large part of the computation of many big social media %software
platforms. Traditionally, most efforts in action recognition have focused on improving accuracy by creating larger architectures. These architectures take as input either a frame or a set of frames (also called a {\em clip}) and produce a prediction. These predictions are then aggregated over time. The frames or clips are either sampled densely \cite{simonyan2014two,yue2015beyond} or randomly \cite{wang2016temporal}. 

Videos, however, provide an opportunity for reducing computational cost in multiple ways. First, videos contain highly temporally redundant data, making it easier to skip parts without losing much information \cite{fan2018watching}. Second, some parts of a video can be more discriminative than others, due to their content, or other phenomena like blur, occlusions, etc. Supporting this intuition, \citet{whatmakes} %Huang et al.\citet{whatmakes} 
show experimentally that using an oracle to make an optimal selection of frames (or clips), produces more accurate classification results than using the entire video. Additionally, \citet{onlytime} show %experimental evidence 
that many action classes in standard datasets do not require motion or temporal information to be identified. For a human observer, a few still frames are often discriminative enough. This suggests that large parts of a video can be discarded. 

Several recent works\cite{korbar2019scsampler,wu2019adaframe,faster2020} have successfully leveraged these principles to reduce computational cost at test time. These methods have used a common strategy: they use an inexpensive way to decide which regions of the video are important and discriminative, and only process those with an expensive method. This general problem has been referred to as frame or clip selection.  
%\marcus{Our idea seems to be somewhat independent of trimmed vs. untrimmed videos?}
%\marcus{Also we compare to other methods in the end with our method!?}
%\marcus{maybe we should discuss short & information rich videos clips vs. long sparse videos, where sub-selection is much easier, i.e. not talk about trimmed vs. untrimmed?}
While very successful, most frame and clip selection methods have focused on a particular domain of action recognition, namely long, and frequently sparse videos with a typical length of a minute or more, %\marcus{please check if that is correct} %which is untrimmed videos\marcus{maybe we have to say what is meant by "trimmed videos" and also why it is "interesting". reading this, it seems a bit like "people have focused on the more general problem, and we focus on something less exciting/interesting, as we need "trimmed video""}. 
%Thus, they present results using 
e.g. ActivityNet\cite{caba2015activitynet}, Sports1M\cite{karpathy2014large}, FCVID\cite{jiang2017exploiting}, Youtube 8M\cite{abu2016youtube}. This is indeed the domain where discarding portions of a video is easier and has potentially the largest effect. In contrast, the problem of frame selection in short videos of a few seconds remains much less explored, probably due to its difficulty. 

In this paper we propose a method to do frame selection in the core, standard activity classification setting of trimmed video clips. Part of the challenge in this setting is that ``good" frames are often temporally close together  within a video. Since most existing frame selection methods consider the value of choosing a frame one at a time, the selected frames tend to only represent part of the action. In other words, the diversity of frames, and their ability to tell a story are disregarded. We also show that using language features along with visual features helps improve the performance.

%\marcus{I would focus on this aspect: }
To handle these challenges, we propose a model that, in addition to considering the discriminative value of a single frame, also considers its relation to others in a video. We do this by using an attention and a relational network \cite{meng2019frame,sung2018learning}, that examines the value of frames jointly. We learn our {\bf S}ampling through {\bf M}ulti-frame {\bf A}ttention and {\bf R}elations in {\bf T}ime, which we dub the {\em \alg{}}selection network. %\LS{not sure if this is a bit obnoxious for a name... I'll keep brainstorming}. 

We test our \alg frame selection network on several trimmed action recognition datasets, including Something-something, UCF101 and subsets of Kinetics. We observe that in all of them the proposed method outperforms the baselines, including using the full video, while reducing the computational cost by a factor of 4 to 10, depending on the dataset. We also test the proposed method on the untrimmed setting in ActivityNet and FCVID, where we get higher accuracies than all previous work on frame selection. Further, we extend our frame selection approach to select frames that are then passed at test time to deep action recognition models and show that we obtain state-of-the-art results on UCF101 and HMDB51 which are trimmed video datasets, showing that frame selection can be an important step to improve accuracy in trimmed action recognition.

\section{Related Work}

The field of action recognition is wide, and includes a large variety of subproblems, and families of methods. Here we focus on the two areas within action recognition that are most relevant to our work: frame selection as well as attention and relational models.

\noindent\textbf{Frame Selection.}
Selecting important frames for action recognition is a relatively new area. Many approaches have successfully trained a reinforcement learning (RL) agent approach that examines one frame at a time, to predict how many frames can be skipped. 

AdaFrame\cite{wu2019adaframe} leverages RL, in combination with an LSTM that is augmented with memory that helps providing context information for selecting frames to use. Given a frame, it generates a prediction of the action class and it decides which frame to observe next and computes the expected reward of seeing more frames. 
FastForward\cite{fan2018watching} is an end-to-end reinforcement learning approach. It consists of two sub networks: an adaptive stop network and fast forward network. The adaptive stop network can either let the frame sampling continue or stop. The fast forward network has a set of several actions (going backwards or going forward with varying seconds). The RL agent learns to skim through the video.

FrameGlimpse\cite{yeung2016end} follows the intuition that detecting an action is dependent on observation and refinement. Based on this, FrameGlimpse relies on a recurrent neural network (RNN) based agent that observes and decides where to look next. Given the current frame, the agent also decides whether to emit a prediction based on a confidence score. If the agent is not confident enough then it decides to look ahead. 

Multi-agent Reinforcement Learning (MARL)\cite{wu2019multi} formulates the frame sampling procedure as multiple parallel Markov decision processes which aim at picking frames by gradually adjusting an initial sampling. They have a context-aware observation network which jointly models context information among nearby agents and historical states of a specific agent. They also have a policy network which generates a probability distribution over a predefined action space.

SCSampler\cite{korbar2019scsampler} is a lightweight clip-sampler that can efficiently obtain the most salient temporal clips within a long video. They sample features directly from compressed videos and also from the audio obtained from the video. Attention aware sampling (AAS) \cite{dong2019attention} uses an agent which discards irrelevant frames using attention. They consider the frame selection procedure as a Markov decision process and train an agent without extra labels through deep reinforcement learning.

While all these approaches showed great results, they have mostly focused on the scenario of untrimmed videos. SCSampler does however report results on Kinetics\cite{carreira2017quo}, however, it  requires audio as an extra modality. Untrimmed videos contain significant parts of unnecessary data and discarding them is easier than discarding frames from trimmed videos. In contrast to previous work, we propose a method that instead of selecting frames by considering one at a time,  considers  them  jointly.

\noindent\textbf{Attention and Relational Models.} The concept of attention was introduced by \citet{bahdanau2014neural} for the objective of machine translation. This concept of attention is based on the concept that the neural network will learn how relevant different samples are regarding the desired output state in a sequence, or image regions. These values of importance are specified as weights of attention and are generally calculated at the same time as other model parameters trained for a specific goal. 

Attention has been used in first person action recognition by having a joint learning of gaze and actions \cite{li2018eye}, by using object-centric attention \cite{sudhakaran2018attention} or via event modulated attention \cite{shen2018egocentric}. The use of attention to weigh spatial regions representative of a particular task was done by generating spatial attention masks implicitly by training the network with video labels \cite{sharma2015action,zhang2018adding,girdhar2017attentional}. Temporal attention was used for action recognition by detecting change in gaze \cite{piergiovanni2017learning,shen2018egocentric}. 

LRCN \cite{donahue16pami} introduced a simple LSTMs for frame-aggregation across time for action recognition. Non-local Networks \cite{wang2018non} introduce a residual self attention block in convolutional networks to aggregate information across all temporal and/or spatial locations. 

Inspired by the relation-net\cite{sung2018learning}, relation attention was proposed to deal with the task of facial emotion recognition \cite{meng2019frame}. They believed that having a global level representation of features in addition to the local level representation helps obtain better results. We improve upon this approach by adding relation-temporal attention to add a global representation to our temporal attention.

\section{\alg Frame Selection}
%\marcus{we need to start with a higher level intro.}

The proposed approach is designed to use a small portion of the overall computational cost in selecting the best frames. These frames will then be classified using a more computationally expensive model. Therefore, we use a very lightweight representation of the frames as input to the \alg frame selection model. 

\begin{figure}[t]
\centering
\includegraphics[width=\linewidth]{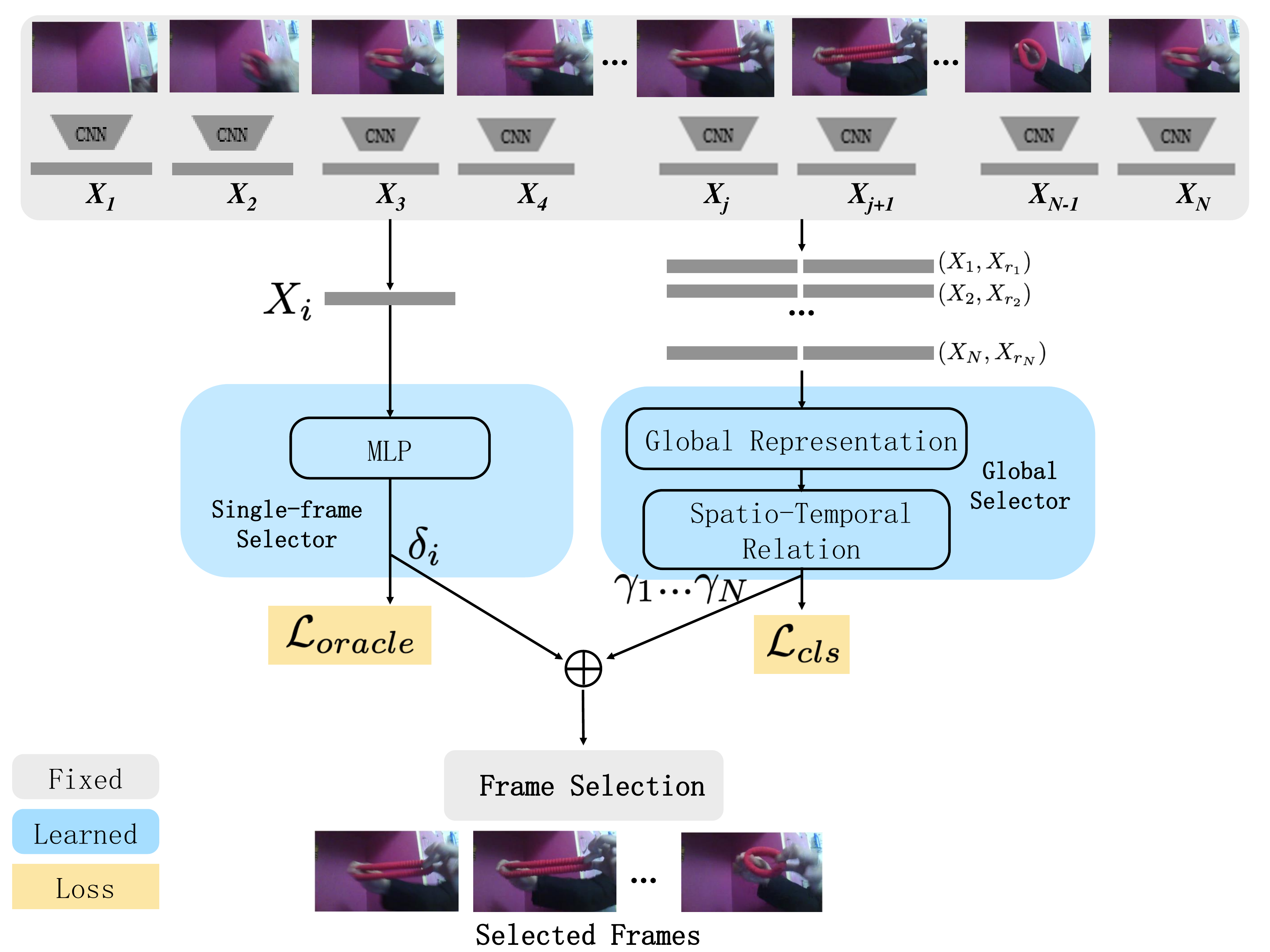}
\caption{Overview of the \alg frame selection. } 
\label{fig:overview}
\end{figure}

The model consists of two streams. The first considers the information of the frames one at a time, and outputs a score $\delta_i$ for each frame, which represents how useful the frame is for classification. The second stream considers the entire video at a time. It takes as input pairs of frames, and uses an attention and relational network to also obtain a score $\gamma_i$ of how useful these pairs of frames are. Both scores are then multiplied, to obtain a final score of how good each frame is. Given a budget of $n$ frames, we now select the top $n$ frames with the highest discriminative score, and use an expensive, high quality classifier for the final prediction. An overview of the method can be seen in Fig.~\ref{fig:overview}. We now describe each of the components in detail.

% \begin{figure*}[t]
% \centering
% \includegraphics[width=0.7\textwidth]{./figures/overview_2.pdf}
% \caption{Overview of the \alg frame selection.}
% \label{fig:overview}
% \end{figure*}

\subsection{Feature Representation}

We choose the lightweight  MobileNet\cite{sandler2018mobilenetv2} to extract the visual features of each frame, to minimize the computational cost of this stage. We also make the observation that, in addition to the visual features, we can use language features associated with the content of the frame. The intuition behind this is to enrich the representation with terms that are related to the content of the image. One could imagine that if for an action class like ``kayaking", having associated words like {\em water}, {\em boat}, or {\em paddle} can help discrimination in cases where the kayak is not as apparent visually.  
%We compute the language features by taking the vector of predictions, choosing the first 10 most probable classes. % and storing it in a one-hot vector using a word embedding matrix vocab. 
We run Mobilenet pre-trained on Imagenet on the frames, and take the top 10 Imagenet classes with highest probability. The names of these classes are then embedded with a pre-trained GloVe \cite{pennington2014glove} over Wikipedia 2014 and average over the 10 classes.
%do a dot product between the 2 vectors to obtain a language embedding which is then concatenated with the 
The language embedding is then concatenated with the visual features resulting in a feature vector $X_i$ for each frame $i$.
%We use a pre-trained GloVe \cite{pennington2014glove} over Wikipedia 2014 and do a dot product between the 2 vectors to obtain a language embedding which is then concatenated with the visual features to have a set of concatenated feature vectors $X_i$ for each frame $i$.

\subsection{Single-frame Selector}

This stream is designed to be extremely fast. We build on the observation from \citet{whatmakes} that an oracle that looks at the predictions from an expensive network, and selects the frames with the highest confidence for the ground truth class, actually outperforms using the entire video for prediction. Thus, we use a simple multi-layer perceptron (MLP) that takes as input a feature vector $X_i$, and computes the confidence of the classification for the ground truth. This MLP has 2 layers, and is trained using the oracle mentioned before wherein each frame outputs the probability of that frame with respect to the ground truth class. At training time we can obtain the ground truth probability of each frame using an expensive model trained on the dataset we are looking at. The model is trained on that. At test time $\delta_i$ is predicted by the trained model as the importance score of a particular frame.  %Overall, this module aims to minimize the loss $\mathcal{L}_{oracle}$ that is described in Eq.~\ref{eq:l_oracle}, given ground truth labels  $\hat{y}$.

%\begin{equation}
 %  \mathcal{L}_{oracle}= -\sum_{i=1}^{N}\hat{y}\log(p_{i}(\hat{y}))
 %   \label{eq:l_oracle}
%\end{equation}

%\noindent\textbf{Network architecture and training details}
% This section details our approach along with the architecture of our model. Given a set of frames, we extract visual features from them by using a ResNet-101 \cite{he2016deep}.

\subsection{Global Selector}

The multi-frame discriminator is designed to use information across frames for selection. This is done by first obtaining a global representation of the video using an attention model over the entire video. Given this global representation, the temporal relationships across frames are learned %spatially \marcus{this also temporally, or?} 
using a relation model and  a long short-term memory (LSTM) network. While lightweight and easy to learn, this network provides information about how useful frames are when considered globally. The global selector uses a relational model to learn temporal relationships across frames over the entire video. This produces an inexpensive global representation of the video.

\noindent\textbf{Pairs of frames.} Consider an input sequence $X = ( X_1,..., X_N), X_i$ represents the concatenated visual and categorical features in  frame $i$ and $N$ represents the total number of frames.
For each frame, we concatenate  a second, randomly selected frame, $X^i_{r},r\in\{1,...,N\}$. The random frame is always chosen from the subsequent set of frames to capture the temporal changes that occur in actions. Some actions will be most recognizable when these pair of frames are only a few frames apart, while others will be more recognizable when they are further apart. This random choice allows the model to be flexible and capture the temporal changes in different classes. The input to the attention model is the concatenation of both vectors  $Z_i=[X_i:X^i_r]$. The output of the network are a set of temporal relation-attention weights $\gamma_1, \gamma_2,.., \gamma_N$. This helps our model to obtain temporal information.

\noindent\textbf{Attention Module.} The coarse self-attention weights $\alpha_i$ are first calculated using a fully connected layer and a sigmoid function \cite{meng2019frame}. The mathematical representation is in Eq.~\ref{eq:self_attention}, where U are network parameters. We now aggregate the input features using these self-attention weights. We do this in order to obtain a global representation $Z'$ of the frame features, as in Eq.~\ref{eq:self_attention}. 

Self-attention weights are learned using individual frames with the help of non-linear mapping. To obtain a more reliable form of attention, we need both local and global features to be used. $Z'$ is aggregated from all local features and hence contains the global information of the video. %some part of every frame in the video. 
Hence, by using $Z'$ we can further refine the attention weights by modeling the relationship between local frame features and $Z'$.

\begin{equation}
    \mathrm{\alpha_{i}=\sigma(Z_{i}U)} \quad and\quad 
        \mathrm{Z'=\frac{\sum_{i=1}^{N}\alpha _{i}Z_{i}}{\sum_{i=1}^{N}\alpha _{i}}}
    \label{eq:self_attention}
\end{equation}

\noindent\textbf{Relation Module.} We can add a sample concatenation and another fully connected layer \cite{sung2018learning} to estimate a relation-attention weight $\beta$. $\Theta_1$ is a parameter of the fully connected layer and $\sigma$ represents the sigmoid function. Using this we have obtained frame attention weights. However, we also want temporal attention weights. We use an LSTM to capture sequential per frame changes. The input to the LSTM at each time step is the dynamic weighted sum using the relational self-attention weights '$\omega_t$'. This is represented in Eq.~\ref{eq:global_rep}.

\begin{equation}
    \mathrm{\beta _{i}=\sigma ([Z_{i}:Z']^{T}\Theta_1)}
\quad and \quad
    \mathrm{\omega_{t}=\sum_{i=1}^{t}\beta _{i}Z_{i}}
    \label{eq:global_rep}
\end{equation}

The temporal attention weights are then calculated as shown in Eq.~\ref{eq:temp_attention} and Eq.~\ref{eq:temp_att}. It is dependent on the previous time step output of the LSTM and the input at that time step. $b$ is a bias vector.

\begin{equation}
    \mathrm{h_t,m_t =LSTM(\omega_{t},h_{t-1}, m_{t-1})}
    \label{eq:temp_attention}
\end{equation}

\begin{equation}
    \mathrm{\lambda _{t}=softmax(Vh_{t}+b)}
    \label{eq:temp_att}
\end{equation}

To compute the relational-temporal weights, we follow the procedure used to obtain relational-frame attention weights, as in Eq.~\ref{eq:gamma}. Here $\Theta_2$ is simply a network parameter.

\begin{equation}
    \mathrm{Z''=\frac{\sum_{t=1}^{N}\lambda _{t}\omega_{t}}{\sum_{t=1}^{N}\lambda _{t}}}
 \quad and \quad
    \mathrm{\gamma _{t}=\sigma ([\omega_{t}:Z'']^{T}\Theta_2)}
    \label{eq:gamma}
\end{equation}

\begin{figure}[t]
\centering
\includegraphics[width=\linewidth]{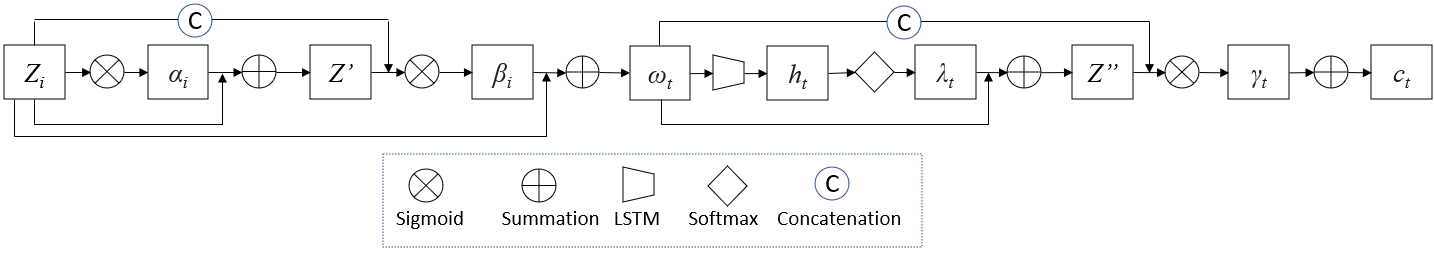}
\caption{Steps involved in calculating the attention weights and intermediaries involved.}
\label{fig:gamma}
\end{figure}

Using these $\gamma_{t}$ we can obtain an attended content vector $c_{t}$ at time 't' using Eq.~\ref{eq:lstm}. Here $h_{i}$ refers to the hidden state of the LSTM at $i$. For classification, $c_{t}$ is fed into an MLP to generate the predicted label $y$. Overall, this module aims to minimize the loss $\mathcal{L}_{cls}$ that is described in Eq.~\ref{eq:cls_loss}, given ground truth labels  $\hat{y}_{t}$. Steps to calculate all the attention weights and intermediaries can be seen in Figure~\ref{fig:gamma}.
\begin{equation}
    c_{t}=\sum_{i=1}^{t}\gamma _{i}h_{i}
    \label{eq:lstm}
\end{equation}
\begin{equation}
    \mathcal{L}_{cls}=-\sum_{i=1}^{C}\hat{y}_{i}\log(y_{i})+\varepsilon \sum_i \sum_j \Theta ^{2}_{i,j}
    \label{eq:cls_loss}
\end{equation}

\section{Experimental Analysis}
In this section we describe all experiments that we conduct to test the behavior of the proposed \alg frame selection network. In the qualitative results analysis subsection we describe the ablation experiments that led to the specific design of the network, justify each of the components and measure their impact. We analyze the behavior of the frame selector components individually (single frame and global selection). We show the generality of the proposed method on several datasets. Finally, we compare to other state-of-the-art frame sampling methods in the untrimmed setting, showing that the proposed method still produces higher accuracy.

\subsection{Experimental Setup}

\noindent\textbf{Datasets.} We use 6 of the most popular benchmark datasets throughout our experimental analysis. 
We use the Something-something-v2 dataset \cite{goyal2017something} for our extensive ablation study. The purpose of the ablation study is to drive the design choices through experimental evidence. We choose this for the ablation study because we think that it contains the types of actions where relations of frames over time matter more. This allows us to truly evaluate the effect of the global model that we propose. In particular, the action classes in this dataset are designed to focus on the action, (eg.: ``put something") instead of on an object (eg.: ``playing guitar"). As a result, actions tend to have more temporal structure, and relations across frames may matter more. 

The Something-something dataset has a total of 168,913 training videos and 24,777 validation videos with a total of 174 classes.
After the ablation study, we show the generality of our approach by testing it in other datasets as well. The Kinetics\cite{carreira2017quo} dataset is one of the most widely used large-scale datasets in action recognition. %Since the goal of this experiment is to measure the gain of the proposed method under as many different types of videos as possible, 

We use two subsets \cite{onlytime} of Kinetics that have been identified as containing mostly temporal information and mostly static information. In our experiments we refer to these as Kinetics-Temporal and Kinetics-Static. These subsets were created using a human perceptual test, where users are asked to identify the class of a video where the frames are not in order, therefore removing temporal information. Static classes are those that users could identify without temporal information. Temporal classes are those that users were not able to identify when the frames were not in order.% These two subsets provide an interesting setting for experimentation, because the statistics of the data (eg.: size of videos, frame rate, platform that the videos came from, etc) are common for both datasets, but the nature of the action classes differ. 
Each of the two splits contains 32 classes. The temporal subset consists of 26509 videos and the static subset consists of 23675.
For our generality tests, we also use the well-known UCF101\cite{soomro2012ucf101} dataset which contains 101 classes and about 13K videos. %This was one of the first widely used action classification datasets and thus we think that it can give additional useful reference for comparison. 

We also extend our approach as a pre-processing step for more complex models and compare performances on HMDB51 which contains 51 classes and 6849 video clips along with UCF101. 
Previous frame selection for action recognition have focused on untrimmed videos. %, as we mentioned before. Therefore, they have not tested on any of these datasets and have not released code so that we could test the methods on these datasets ourselves. 
In order to compare with them, we use ActivityNet\cite{caba2015activitynet} and FCVID\cite{fcvid}. ActivityNet consists of 19994 videos, and contains 200 classes. As the testing labels are not available publicly, the reported performances are on the validation set. FCVID is made up of 91, 223 videos taken from YouTube having an average duration of 167 seconds, and these are annotated into 239 classes.

\noindent\textbf{Implementation Details.} 
%For selecting frames, we use visual features obtained from MobileNet v2 \cite{sandler2018mobilenetv2}. Taking the 1000 class predicted vector, we then choose first 10 most probable classes and store in a vector. This is then converted to a one-hot encoding vector using a word embedding matrix vocab. We use pretrained GloVe \cite{pennington2014glove} over wikipedia 2014 and do a dot product between the 2 vectors to obtain a language embedding which is then concatenated with the visual features to have a set of concatenated feature vectors. We show in our ablation study that having a semantic embedding helps our frame selector to make better informed decisions.
As mentioned before, the lightweight features used for frame selection are computed using MobileNet\cite{sandler2018mobilenetv2} and GloVe\cite{pennington2014glove}. After the frame selection is done, we can use a more expensive and high-quality feature representation. In our experiments, we use three different backbones: ResNet-152, ResNet-101\cite{he2016deep}, and Inception-v3\cite{szegedy2017inception}. The backbones are pre-trained either on ImageNet\cite{deng2009imagenet} or Kinetics. These architectures are representative of the state-of-the-art, and are chosen according to what other methods that we want to compare to have used. %While having multiple backbones and pre-training datasets makes experiments more cumbersome, it allows us to simply swap the frame selection component and directly observe the effect on the accuracy without any other confounders. 

We use Pytorch for implementation. All frames are resized to 224x224. We use mini-batch stochastic gradient descent, with a momentum of 0.9. We run 200 epochs on UCF101 and the Kinetics subsets, and 100 epochs on Something-something dataset and Activitynet due to the computational requirements for these larger scale datasets. We use a batch size of 128 for UCF101 and the Kinetics subsets and a batch size of 64 for the Activitynet and something-something datasets. The initial learning rate is set at 0.0001 and reduces by 10 after every 25 epochs.   %We use a 1-layer LSTM with 1024 hidden units for the temporal attention.\\

\noindent\textbf{Baselines.} We compare the performance of our frame selection model with that of random and uniform frame selection. Random frame selection picks frames uniformly at random from the entire video, while uniform frame selection picks frames that are evenly spaced. Once the frames are picked, we predict an action by average pooling the predictions of every selected frame using one of the expensive backbones. In addition to these baselines, we compare to other state-of-the-art frame sampling methods, including Adaframe\cite{wu2019adaframe}, FastForward\cite{fan2018watching}, FrameGlimpse\cite{yeung2016end} and MARL\cite{wu2019multi}. 

\subsection{Ablation Study on the \alg Frame Selection}
\label{subsec:ablation}

Here, we look at the impact of the feature representation (visual and categorical), the choice of frame selector (the global multi-frame selector and the single-frame discriminator), and the use of pairs of frames. We use the Something-something-v2 \cite{goyal2017something} dataset for this study. Table~\ref{table:ablation} shows the results. 

We first test and compare the use of the simple visual features (from MobileNet), then combining them with the categorical ones (from GloVe). We use the global selector for this initial test. We observe that the addition of semantic language features helps, supporting the intuition that using words related to the content of a frame actually helps in the context of frame selection. 
Using that, we examine the effect of different selectors: the single-frame selector, the global selector, and the combination of both. We observe that the combination of both is the best choice, suggesting that these two selectors behave in different but complementary ways. %We will analyze this behavior further in the following section. 

We also measure the impact of using pairs of frames as input to the global selector. While we use a relational component inside the selector, adding pairs would give an additional mechanism to consider frames jointly. We observe that this does indeed help.  Since we use random frames, we report the average accuracy in Table~\ref{table:ablation}. The standard deviation on the Something-something-v2 \cite{goyal2017something} dataset on 10 random runs was 0.067 using Inception v3 and 0.082 using Resnet-152.

%With this we conclude the design of the \alg frame selection method. 

\setlength{\tabcolsep}{3pt}
\begin{table}[tb]
\centering
\caption{Ablation study to determine the effect of each of the components of the \alg frame selection network. In the table, the best configuration of each section is the setting used for the section below. We use 26 frames for all experiments. Something-something-v2 dataset. 'G' represents GLOPs, 'VF' represents standalone visual features, 'SFS' and 'GS' stand for single frame selection and global selection respectively.}% The prefix 'I' and 'R' represent backbones as Inception v3 and Resnet-152 respectively }
\label{table:ablation}
\begin{tabular}{llrrrr}
%\hline\noalign{\smallskip}
\toprule
&&\multicolumn{2}{c}{Inc v3} & \multicolumn{2}{c}{Res-152}\\
& Method & Acc & G & Acc & G\\
\noalign{\smallskip}
\cmidrule(lr){1-2}\cmidrule(lr){3-4}\cmidrule(lr){5-6}
\noalign{\smallskip}
\multirow{3}{*}{Baselines} & Random & 44.2 & 152 & 45.8 & 277\\ 
& Uniform & 49.6 & 152 & 50.8 & 277\\ 
& All frames & 58.8 & 607 & 60.1 & 1105\\
\cmidrule(lr){1-2}\cmidrule(lr){3-4}\cmidrule(lr){5-6}
\multirow{2}{*}{\specialcell{Input \\ Features}} & VF  & 58.3 & 182 & 60.2 & 308\\
& VF + GloVe & 59.2  & 183 & 60.3  & 309\\
\cmidrule(lr){1-2}\cmidrule(lr){3-4}\cmidrule(lr){5-6}
\multirow{3}{*}{Selector} & SFS only & 59.7 & 155& 60.7 & 279 \\
& GS only & 59.2 & 183 & 60.3 & 309\\
& SFS + GS & 60.6 & 184& 61.0 & 310\\
\cmidrule(lr){1-2}\cmidrule(lr){3-4}\cmidrule(lr){5-6}
{\bf \alg} & 2-frame input & {\bf 60.8} & 186 &{\bf 61.2} & 311\\
\bottomrule
%\hline
%\multirow{3}{*}{Baselines} & Random & Res-152  \\ 
%& Uniform & Res-152 & 50.8 & 277\\ 
%& Full Video & Res-152 & 60.1 & 1105\\
%\hline
%\multirow{2}{*}{\specialcell{Input \\ Features}} & Visual Features & Res-152 & 58.3 & 308\\
%& Visual Features + GloVe & Res-152 & 59.2  & 309\\
%\hline
%\multirow{3}{*}{Selector} & Single-frame Selection only & Res-152 & 60.7 & 279 \\
%& Global Selection only & Res-152 & 60.3 & 309 \\
%& Single-frame + Global Selection & Res-152 & 61.0 & 310\\
%\hline
%{\bf \alg} & 2-frame input &  Res-152 & {\bf 61.2} & 310\\
%\hline
\end{tabular}
\end{table}
\setlength{\tabcolsep}{1.0pt}

\subsection{Analysis of the Behavior of \alg Frame Selection} 
\label{subsec:analysis}

\begin{figure}[tb]
\centering
\includegraphics[width=\linewidth]{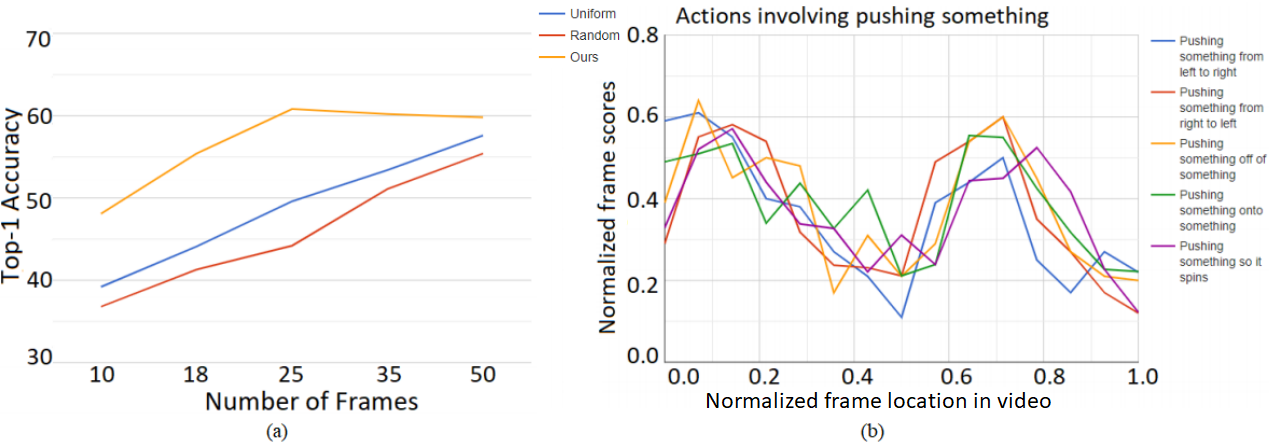}
\caption{(a) Behavior of different sampling strategies with respect to number of frames. Orange represents \alg, blue represents uniform selection and red represents random selection (b) Comparison of the importance score of semantically similar actions. We can see a striking resemblance for all actions involving pushing.}
\label{fig:semanticsimilar}
\end{figure}

\noindent\textbf{Number of Selected Frames. } First, we measure the impact of selecting different number of frames. For this, we vary the number of selected frames between 10 and 50, and measure the impact on accuracy and GFLOPs and compare with random and uniform sampling. The results are in Fig.~\ref{fig:semanticsimilar}(a). We choose the Something-something dataset, and the Inception-v3 as backbone. We see that as the number of frames increases, the uniform and random frame selection perform strictly. The proposed method performs much better than these baselines across frames. It is also interesting that the accuracy increases and reaches a peak, and then slowly drops in performance. This behavior confirms the intuition that there is a sweet spot in the number of frames, and that using more than that, will include frames that are harder to classify, which will pollute the prediction.

\noindent\textbf{Frame Selection Across Similar Classes}. We now plot the combined frame score from both selectors, to analyze its behavior. We plot the frame score of classes that are semantically related, to compare if frames scores are also similar. The Something-something dataset contains groups of classes that are very related. We sample 25 videos within a class, for 5 classes and average the importance scores. The plots are shown in Fig.~\ref{fig:semanticsimilar}. 
We see a strong resemblance for actions involving ``pushing", suggesting that the general structure of the action has been captured by the model. %We also compare actions involving ``throwing", which show peaks for frame importance at similar time intervals. It is also interesting to look at the frames selected for one of the classes, in Fig.~\ref{fig:selected_frames}. Indeed, the few selected frames do tell the story of the action. 

%\begin{figure}[t]
%\centering
%\includegraphics[width=\linewidth]{./figures/example_pushing.png}
%\caption{Examples of frames not selected (top) and selected (bottom) for the class ``pushing something from left to right".}
%\label{fig:selected_frames}
%\end{figure}

\begin{figure}[t]
\centering
\includegraphics[width=\linewidth]{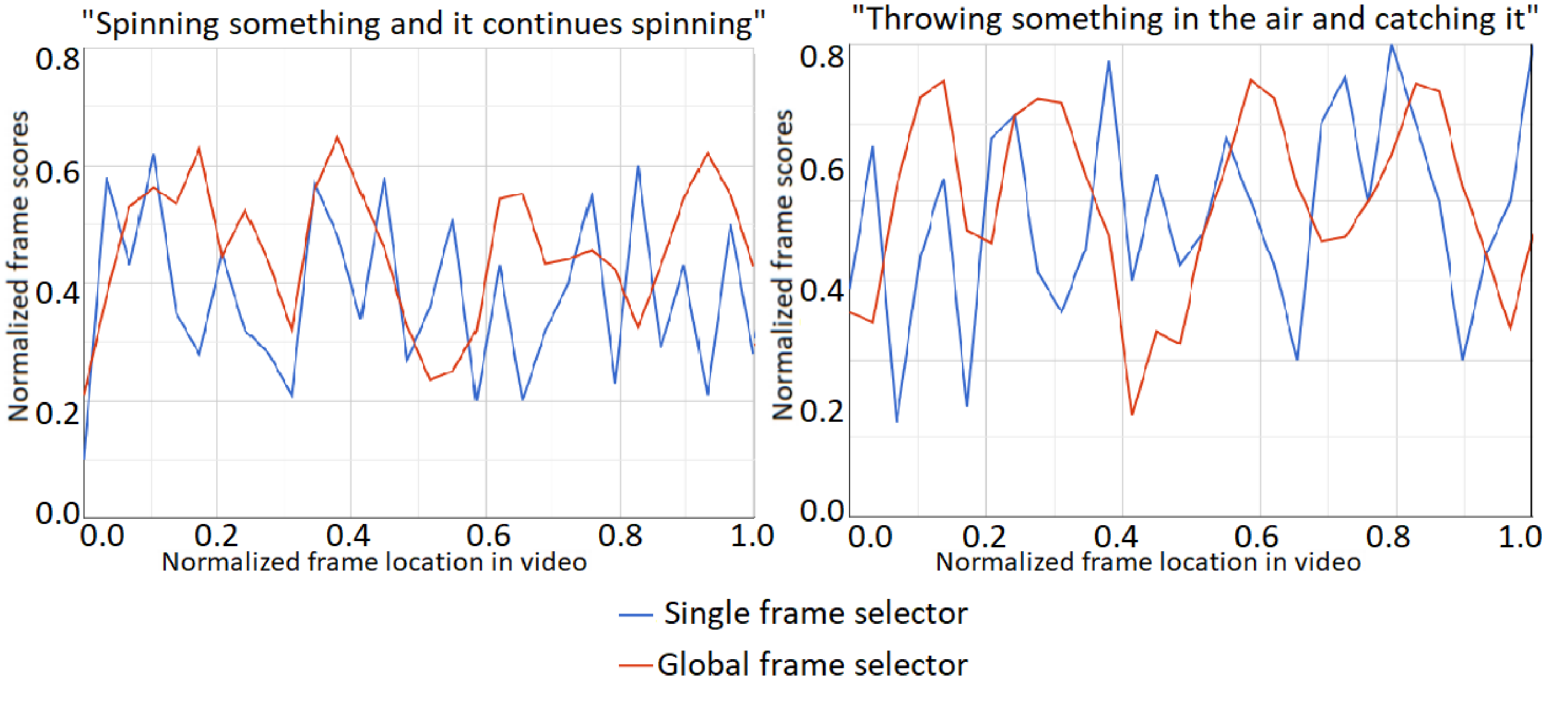}
\caption{Graphical comparison of how the two selection modules give importance scores. We can see that each selector is giving different importance weights to different parts of the video.}
\label{fig:singlevsdouble}
\end{figure}

\noindent\textbf{Selecting Frames with the Global Selector vs. the Single-frame Selector}. We measure whether the pattern of frame selection from the global selector tends to be different from the pattern from the single-frame selector. For this, we randomly sample 25 videos within a class, and score each of their frames with the two selectors. We plot the average score at each frame, in Fig.~\ref{fig:singlevsdouble}. Again we use the Something-something dataset and Inception-v3. While the scores from the single-frame selector change more erratically, the score from the global selector seems to be more temporally consistent. This suggests that frames scores from the global selector are actually more structured.

\begin{figure}[t]
\centering
\includegraphics[width=\linewidth]{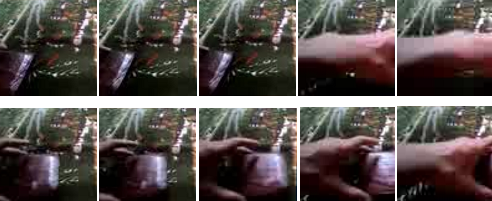}
\caption{Examples of frames not selected (top) and selected (bottom) for the class ``pushing something from left to right". Frames from \cite{goyal2017something}.}
\label{fig:selected_frames}
\end{figure}

\begin{figure}[t]
\centering
\includegraphics[width=\linewidth]{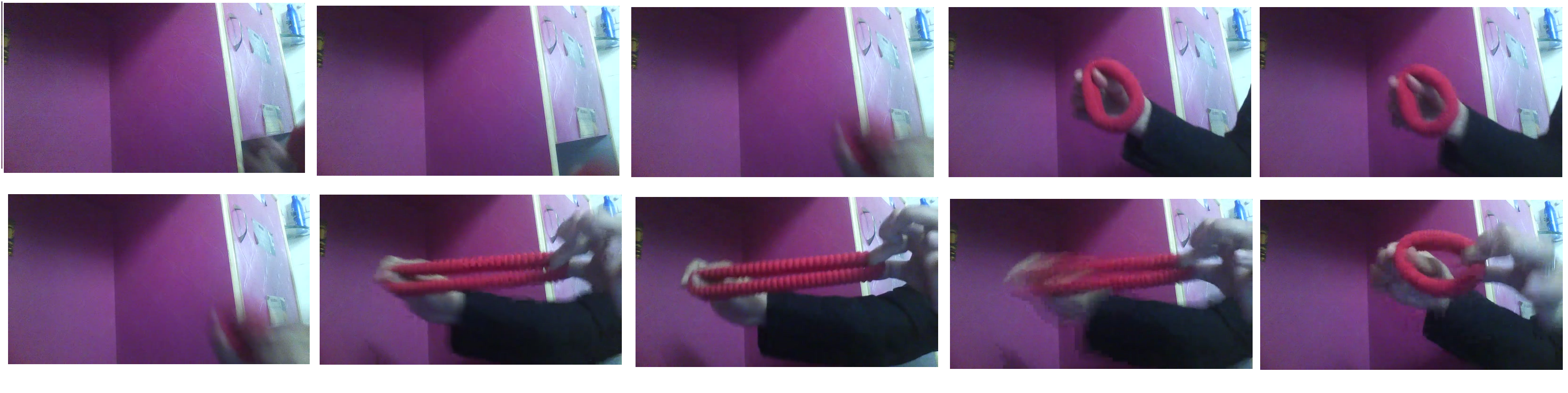}
\caption{Examples of frames not selected (top) and selected (bottom) for the class ``pulling something so that it gets stretched". Frames from \cite{goyal2017something}.}
\label{fig:selected_frames_1}
\end{figure}

\noindent\textbf{Selected frames}. It is also interesting to look at the frames selected for one of the classes, in Fig.~\ref{fig:selected_frames}. Indeed, the few selected frames do tell the story of the action. The class is ``pushing something from left to right". Fig.~\ref{fig:selected_frames_1} shows another example of selected frames.

\section{Quantitative results analysis}

\subsection{Generality of \alg on Additional Datasets}
\label{subsec:other_datasets}

\noindent\textbf{UCF101.} Results for UCF101 can be seen in Table~\ref{table:ucf}. As in the Something-something dataset we observe that the \alg selection outperforms the baselines of random and uniform, regardless of the number of frames. We also see that it outperforms using the full video (for all except for using 10 frames) while the ``sweet spot" of number of frames is slightly larger than in the Something-something dataset. This is consistent with the fact that videos in UCF101 are about 7 seconds long, while Something-something are closer to 3 seconds. Therefore it makes sense that the proportion of ``good frames" stays the same. % It is notable that when using 25 frames, the accuracy of the \alg is higher than using the full video, while the amount of GFLOPs is 6 times smaller. \\

% \setlength{\tabcolsep}{4pt}
% \begin{table}[htb]
% \begin{center}
% \caption{The results of various frame sampling techniques on UCF101 using ResNet-152 as the backbone network. The methods names stand for Random, Uniform, Full Video and \alg. }
% \label{table:ucf}
% \begin{tabular}{cccc}
% \hline %\noalign{\smallskip}
% Method & Frames & UCF101 & GFLOPs\\
% \noalign{\smallskip}
% \hline
% \noalign{\smallskip}
% R/U/Full/\alg & 10 &
% 63.2/63.8/{\bf 74.6}/72.8 & 110/110/1969/164 \\
% R/U/Full/\alg & 26 &
% 68.3/69.1/74.6/{\bf 75.3} & 277/277/1969/331 \\
% R/U/Full/\alg & 50 &
% 70.2/70.7/74.6/{\bf 75.5} & 652/652/1969/706 \\
% \hline
% \end{tabular}
% \end{center}
% \end{table}
%\setlength{\tabcolsep}{1.4pt}

%\setlength{\tabcolsep}{4pt}

\begin{table}[t]
\setlength{\tabcolsep}{12pt}
\caption{Baseline frame sampling techniques vs our \alg  with ResNet-152 backbone; \#F: \#frames.}
\label{table:resutls}

\begin{subtable}[t]{0.47\textwidth}
\setlength{\tabcolsep}{2pt}

\begin{tabular}{lrrrrrr}
\toprule
Method & \multicolumn{3}{c}{Accuracy} & \multicolumn{3}{c}{GFLOPs} \\

\midrule
\multicolumn{1}{r}{\emph{\#F}}& \multicolumn{1}{c}{\emph{10}} & \multicolumn{1}{c}{\emph{26}} & \multicolumn{1}{c}{\emph{50}} & \multicolumn{1}{c}{\emph{10}} & \multicolumn{1}{c}{\emph{26}} & \multicolumn{1}{c}{\emph{50}}\\
\cmidrule(lr){2-4}\cmidrule(lr){5-7}
Random & 63.2 & 68.3 & 70.2                      & 110 & 277 & 652\\
Uniform & 63.8 & 69.1 & 70.7                    & 110 & 277 & 652\\
\textbf{\alg} & 72.8 &  {\bf 75.3} &{\bf 75.5}    & 164 & 331 & 706 \\

\midrule
All frames & {\bf 74.6} & 74.6  & 74.6             & 1969 & 1969 & 1969 \\
\bottomrule
\end{tabular}
\caption{UCF101 dataset}
\label{table:ucf}
\end{subtable}
\ \ 
\begin{subtable}[t]{0.51\textwidth}
\setlength{\tabcolsep}{2pt}

%\centering

\begin{tabular}{@{}rrrrrrrrrr@{}}
\toprule
%Method & F & \specialcell{Kinetics\\Temporal} & \specialcell{Kinetics\\Static} & GFLOPs\\
Method & \multicolumn{3}{c}{Temporal, Acc} &
\multicolumn{3}{c}{Static, Acc} &
 \multicolumn{3}{c}{GFLOPs}\\
%\cmidrule(lr){1-3}\cmidrule(lr){4-6}\cmidrule(lr){7-9}
\midrule
 \multicolumn{1}{r}{\emph{\#F}} & \multicolumn{1}{c}{\emph{10}} & \multicolumn{1}{c}{\emph{26}} & \multicolumn{1}{c}{\emph{50}} &
 \multicolumn{1}{c}{\emph{10}} & \multicolumn{1}{c}{\emph{26}} & \multicolumn{1}{c}{\emph{50}} &
 \multicolumn{1}{c}{\emph{10}} & \multicolumn{1}{c}{\emph{26}} & \multicolumn{1}{c}{\emph{50}}\\
 \cmidrule(lr){2-4}\cmidrule(lr){5-7}\cmidrule(lr){8-10}
 Uniform & 59.4 & 60.3 & 62.1      & 60.1 & 60.6 & 61.2           & 110  & 227  & 652\\
 Random & 60.1 & 60.8 & 62.7 &  60.3 & 60.9   &  61.7       & 110  & 227 & 652\\
 SMART & 63.1 & {\bf 64.8}& {\bf 65.4}      & 61.4 & 61.9 &      {\bf 62.6}     &185 & 353 & 728   \\
 \midrule
All frames & {\bf 64.1} & 64.1 & 64.1& {\bf 62.4} & {\bf 62.4} & 62.4    & 2761 & 2761 & 2761  \\
\bottomrule
\end{tabular}
\caption{Kinetics dataset subsets: Temporal and Static}
\label{table:kinetics}
\end{subtable}
\end{table}

% \setlength{\tabcolsep}{4pt}
% \begin{table}[htb]
% \begin{center}
% \caption{ResNet-152, \LS{Shreyank, please include either GFLOPs or time on this table too.}}
% \label{table:hmdbucf}
% \begin{tabular}{ccccc}
% \hline %\noalign{\smallskip}
% Method & N & UCF101 &  \specialcell{Kinetics\\Temporal} & \specialcell{Kinetics\\Static}\\
% \noalign{\smallskip}
% \hline
% \noalign{\smallskip}
% R/U/Full/Ours & 10 &
% 63.2/63.8/{\bf 74.6}/72.8 & 59.4/60.1/{\bf 64.1}/63.1 & 60.1/60.3/{\bf 62.4}/61.4 \\
% R/U/Full/Ours & 25 &
% 68.3/69.1/74.6/{\bf 75.3} 
%  & 
% 60.3/60.8/64.1/{\bf 64.8} & 60.6/60.9/{\bf 62.4}/61.9
% \\
% R/U/Full/Ours & 50 &
% 70.2/70.7/74.6/{\bf 75.5} & 62.1/62.7/64.1/{\bf 65.4} & 61.2/61.7/62.4/{\bf 62.6} \\
% \hline
% \end{tabular}
% \end{center}
% \end{table}
% %\setlength{\tabcolsep}{1.4pt}

\noindent\textbf{Subsets of Kinetics.} We also show results on the subsampled 32 temporal classes of Kinetics and the 32 static classes \cite{onlytime}. These two subsets are described in detail in the Datasets section. Results are shown in Table~\ref{table:kinetics}. We see that the pattern is similar to all other experiments: \alg outperforms the other sampling baselines, and for the optimal number of frames, it outperforms the full video as well. Also the proposed method behaves slightly differently in these two subsets of Kinetics. %In the Kinetics-Static subset, where temporal information is less rich, the improvement is smaller. 
This is consistent with the expected behavior of the proposed method, which considers the entire video globally, and is able to make selections that are more temporally aware. 

%These two subsets provide an interesting setting for experimentation,because the statistics of the data (eg.: size of videos, frame rate, platform that the videos came from, etc) are common for both datasets, but the nature of the action classes differ. Each of the two splits contains 32 classes. %The temporal subset consists of 26509 videos and the static subset consists of 23675.

% \setlength{\tabcolsep}{4pt}
% \begin{table}[htb]
% \centering
% \caption{The results of various frame sampling techniques on the Kinetics subsets using ResNet-152 as the backbone network. The methods names stand for Random, Uniform, Full Video and \alg. Column F stands for number of frames. }
% \label{table:kinetics}
% \begin{tabular}{ccccc}
% \hline %\noalign{\smallskip}
% Method & F & \specialcell{Kinetics\\Temporal} & \specialcell{Kinetics\\Static} & GFLOPs\\
% \noalign{\smallskip}
% \hline
% \noalign{\smallskip}
% R/U/Full/S & 10 & 59.4/60.1/{\bf 64.1}/63.1 & 60.1/60.3/{\bf 62.4}/61.4 & 110/110/2761/185 \\
% R/U/Full/S & 26 & 
% 60.3/60.8/64.1/{\bf 64.8} & 60.6/60.9/{\bf 62.4}/61.9 & 277/277/2761/353 \\
% R/U/Full/S & 50 & 62.1/62.7/64.1/{\bf 65.4} & 61.2/61.7/62.4/{\bf 62.6} & 652/652/2761/728 \\
% \hline
% \end{tabular}
% \end{table}

\subsection{Performance on untrimmed datasets}

Finally, we compare the performance of the proposed method to previous work for untrimmed video. Therefore, we test on the ActivityNet\cite{caba2015activitynet} and the FCVID\cite{fcvid} datasets. We show that using fewer frames than recent approaches such as Adaframe\cite{wu2019adaframe}, FastForward\cite{fan2018watching}, FrameGlimpse\cite{yeung2016end} we can obtain a higher accuracy. However, we access all frames which makes our approach slower than these. We also compare with LiteEval \cite{wu2019liteeval} which is a lightweight action recognition model. We also compare our approach to
Multi-agent Reinforcement Learning (MARL)\cite{wu2019multi} approach and Dynamic Sampling Networks (DSN) \cite{zheng2020dynamic} by using a model pretrained on Kinetics for fair comparison. Table~\ref{table:activitynet} shows the results.

\setlength{\tabcolsep}{2pt}
\begin{table}[t]
\centering
\caption{Results on ActivityNet and FCVID of the \alg frame selection. Compared to recent state-of-the-art methods, the proposed method outperforms their accuracy. \#F: Number of frames, 10c corresponds to 10 clips used instead of frames}
\label{table:activitynet}
\begin{tabular}{lllllll}
\hline\noalign{\smallskip}
&Pre-&Back-&\multicolumn{2}{c}{ActivityNet} & \multicolumn{2}{c}{FCVID}\\
Method & trained & bone & \#F & Acc & \#F & Acc\\
\noalign{\smallskip}
\hline
\noalign{\smallskip}
FastForward & Imagenet & Inc v3 &  9.61 & 58.1 & 15.34 & 73.3\\
FrameGlimpse & Imagenet & VGG16 & 9.42 & 62.8 & 9.26 & 71.7\\
Adaframe &  Imagenet & Res101 & 8.65 & 71.5 & 8.21 & 80.2\\
LiteEval & Imagenet & Res101 & - & 72.7 & - & 80.0\\
%\hline
{\bf SMART} & Imagenet & Res101 & 8 & 71.4 & 8 & 80.8\\
{\bf SMART} & Imagenet & Res101 & 10 & {\bf 73.1 } & 10 & \bf 82.1\\
%Ours + InceptionV3 & Imagenet  & 72.6 & 9\\
%\hline
%\textbf{Ours + ResNet152} & Imagenet & 73.1 & 9\\
\hline
%MARL + ResNet101 & Kinetics & 81.5 & 25\\
%MARL + InceptionV3 & Kinetics  & 82.3 & 25\\
DSN & Kinetics & Res18 & 10c & 68.0 & - & -\\
DSN & Kinetics & Res34 & 10c & 82.6 & - & -\\
MARL  & Kinetics & Res152 & 25 & 83.8 & - & -\\

%\hline
%Ours + ResNet101  & Kinetics & 81.7 & 25\\
%Ours + InceptionV3  & Kinetics & 82.8 & 25\\
\textbf{SMART} & Kinetics & Res152 & 24 & \textbf{84.4} & - & -\\
\hline
\end{tabular}
%\vspace{-1cm}
\end{table}
\setlength{\tabcolsep}{1.4pt}

\subsection{Extension of \alg as a pre-processing step}
We look at the results of using our approach as a pre-processing step to Temporal Segment Networks (TSN) \cite{wang2016temporal} and using the selected frames at inference in Table~\ref{table:extend}. To the best our knowledge this gives us state-of-the-art results on UCF101 and HMDB51. We compare with other recent state-of-the-art approaches such as two-stream networks \cite{simonyan2014two,gowda2017human}, DynaMotion \cite{asghari2020dynamic}, I3D \cite{carreira2017quo} and Knowledge Integration network (KI-Net) \cite{zhang2020knowledge} which are among the latest state-of-the-art approaches. We also add comparison with AAS \cite{dong2019attention} as a frame selection approach.

\setlength{\tabcolsep}{4pt}
\begin{table}[tb]
\begin{center}
\caption{Extending \alg as a pre-processing step to state-of-the-art deep learning approaches. The '+ Kinetics' indicate that the backbone is pre-trained with Kinetics. }
%\marcus{add citation to methods}
%\marcus{are the numbers you list from the corresponding papers or where did you get them from?}
%\marcus{I would do this table more similar to Table 5, listing the backbone as additional pre-trained/backbone as additional column}
\label{table:extend}
\begin{tabular}{lccc}
\hline %\noalign{\smallskip}
Method & Backbone & UCF101 & HMDB51\\
\noalign{\smallskip}
\hline
\noalign{\smallskip}
Two-stream & VGG & 92.5 & 62.4\\
I3D & Inc v3 & 98.0 & 80.7 \\
DynaMotion + I3D & Inc v3 & 98.4 & 84.2 \\
TSN  & BN-Inc & 94.2 & 69.9 \\
KI-Net & Res-152 & 97.8 & 78.2 \\
\hline
AAS  & TSN & 94.6 & 71.2 \\
\textbf{\alg} & TSN  & \textbf{95.8} & \textbf{74.6} \\
\hline

 AAS & TSN+Kinetics & 96.8 & 77.3\\
\textbf{\alg}    & TSN+Kinetics & \textbf{98.6} & \textbf{84.3}\\
\hline
\end{tabular}
\end{center}
\end{table}

\subsection{Improving performances of other models}

Here, we show that using our model to select frames and pass the selected frames at inference helps to improve the performance of models such as I3D \cite{carreira2017quo}, STM-ResNet \cite{feichtenhofer2017spatiotemporal} and ISTPAN \cite{du2018interaction}. This can be seen in Table~\ref{table:extendsupp}.

\setlength{\tabcolsep}{4pt}
\begin{table}[h!]
\begin{center}
\caption{Extending SMART to other approaches }
\label{table:extendsupp}
\begin{tabular}{lccc}
\hline %\noalign{\smallskip}
Method & UCF101 & HMDB51\\
\noalign{\smallskip}
\hline
\noalign{\smallskip}

ISTPAN  &  95.5 & 70.7 \\
ISTPAN + SMART & \textbf{96.4} & \textbf{72.1} \\
\hline
I3D & 98.0 & 80.0 \\
I3D + Smart & \textbf{98.2} & \textbf{81.1} \\
\hline
 STM-Resnet & 94.2 & 68.9\\
 STM-Resnet + SMART & \textbf{94.9} & \textbf{69.7}\\
\hline
\end{tabular}
\end{center}
\end{table}

\section{Conclusion}

We have proposed a method for frame selection in the domain of trimmed videos, that we refer to as \alg frame selection. The method addresses the issue of considering all frames in a video at once, instead of individually, therefore making decisions globally. The proposed method outperforms the accuracy of the baselines on 3 different action classification datasets, while it reduces the computation cost up to 4 times. Further, it outperforms recent frame selection approaches on untrimmed videos in accuracy. Also, it can be extended as a pre-processing step to obtain state-of-the-art accuracy on 2 benchmarks.

\section{Potential Ethical Impact}
This work is about efficient and effective recognition in videos and shares benefits and concerns with other video recognition models. Being able to recognize content in videos more effectively potentially allows positive impact for users when accessing video, e.g. during search. It might also allow to more effectively remove harmful content, although we do not experiment with this kind of data in this work. However, before pursuing any such use cases it is important to analyze the models for potential algorithmic biases, either obtained during training our model or inherited from pre-trained models our approach is using.
\medskip
\small
\bibliography{biblioLong,rohrbach,egbib}

\end{document}